\documentclass[11pt]{article}
\usepackage{acl2016}
\usepackage{times}
\usepackage{latexsym}
\usepackage{graphicx}
\usepackage{amsmath,bm}
\usepackage{amssymb}
\usepackage{url}

\aclfinalcopy 
\def\aclpaperid{600} 

\title{Cross-Lingual Morphological Tagging for Low-Resource Languages}
    
\author{Jan Buys\\
        Department of Computer Science\\
	University of Oxford\\
	{\tt jan.buys@cs.ox.ac.uk}
      \And
        Jan A. Botha\\
	Google Inc.\\
        London\\
	{\tt jabot@google.com}}	  
\date{}

\begin{document}

\maketitle

\begin{abstract}
Morphologically rich languages often lack the annotated linguistic 
resources required 
to develop accurate natural language processing tools.
We propose models suitable for training morphological taggers with rich tagsets
for low-resource languages without using direct supervision.
Our approach extends existing approaches of projecting part-of-speech 
tags across languages, using bitext to infer constraints on the possible
tags for a given word type or token.
We propose a tagging model using Wsabie, a discriminative embedding-based model 
with rank-based learning.
In our evaluation on 11~languages, on average this model performs on par with
a baseline weakly-supervised HMM, while being more scalable.
Multilingual experiments show that the method performs best when
projecting between related language pairs.
Despite the inherently lossy projection, we show that the morphological tags predicted 
by our models improve the downstream performance of a parser by +0.6 LAS on average.
\end{abstract}

\section{Introduction}

Morphologically rich languages pose significant challenges for Natural Language
Processing (NLP) due to data-sparseness caused by large vocabularies. 
Intermediate processing is often required to address the limitations of only 
using surface forms, especially for small datasets. 
Common morphological processing tasks include
segmentation~\cite{CreutzL07,SnyderB08}, paradigm 
learning~\cite{DurrettD13,AhlbergFH15}
and morphological tagging~\cite{MullerS15}.
In this paper we focus on the latter.

Parts-of-speech (POS) tagging is the most common form of syntactic annotation.
However, the granularity of POS varies across languages and annotation-schemas,
and tagsets have often been extended to include tags for morphologically-marked 
properties such as number, case or degree.
To enable cross-lingual learning, a small set of universal
(coarse-grained) POS tags have been proposed~\cite{PetrovDM12}.
For morphological processing this can be complemented with a set 
of attribute-feature values that makes the annotation more 
fine-grained~\cite{Zeman08,SylakGlassmanKYQ15}.

Tagging text with morphologically-enriched labels has been shown to benefit 
downstream tasks such as parsing~\cite{TsarfatyEa10} and semantic role 
labelling~\cite{HajicEa09}.
In generation tasks such as machine translation these tags can help to
generate the right form of a word and to model agreement~\cite{ToutanovaSR08}.
Morphological information can also benefit automatic speech recognition for
low-resource languages~\cite{BesacierBKS14}.

However, annotating sufficient data to learn accurate morphological taggers
is expensive and relies on linguistic expertise, and is therefore
currently only feasible for the world's most widely-used languages.
In this paper we are interested in learning morphological taggers without 
the availability of supervised data. 
A successful paradigm for learning without direct
supervision is to make use of word-aligned parallel text, with a 
resource-rich language on one side and a resource-poor language on
the other side~\cite{YarowskyNW01,FossumA05,DasP11,TackstromDPMN13}.

In this paper we extend these methods, that have mostly been proposed
for universal POS-taggers, to learn weakly-supervised morphological taggers.
Our approach is based on projecting token and type constraints across
parallel text, learning a tagger in a weakly-supervised manner from the
projected constraints~\cite{TackstromDPMN13}.
We propose an embedding-based model trained with the Wsabie 
algorithm~\cite{WestonBU11}, and compare this approach against a baseline
HMM model.

We evaluate the projected tags for a set of languages for which morphological 
tags are available in the Universal Dependency corpora.
To show the feasibility of our approach, and to compare the performance of 
different models, we use English as source language.
Then we perform an evaluation on all language pairs in the set of target
languages which shows that the best performance is obtained when projecting 
between genealogically related languages.

As an extrinsic evaluation of our approach, we show that NLP models can benefit 
from using these induced tags even if they are not as accurate as tags produced 
by supervised models, by evaluating the effect of features
obtained from tags predicted by the induced morphological taggers in dependency parsing.

\section{Universal Morphological Tags}

In order to do cross-lingual learning we require
a common morphological tagset.
To evaluate these models we require datasets in multiple 
languages which have been annotated with such a consistent schema.
The treebanks annotated in the Universal Dependencies (UD) 
project~\cite{MarneffeEa14} are suitable for this purpose.

All the data is annotated with universal POS tags, a set of 17 
tags\footnote{This extends, but is not fully consistent with, the set of 12 
tags proposed by \newcite{PetrovDM12}.}.
We use UD v1.2~\cite{NivreEa15}, which contain 25 languages 
annotated with morphological attributes (called features).
In addition to POS, there are 17 universal \emph{attributes}, which each takes one of a 
set of \emph{values} when annotated.
The morphological tag of a token denotes the union of its morphological 
attribute-value pairs, including its POS.

Although the schema is consistent across languages, there are language-specific
phenomena and considerations that result in some mismatches for a given pair
of languages.
One source of this is that the UD treebanks were
mostly constructed by fully or semi-automatic conversion of existing
treebanks which had used different annotation schemes.
Furthermore, not all the attributes and
values appear in all languages 
(e.g. additional cases in morphologically-rich languages such as Finnish),
and there are still a number of language-specific tags not in the universal schema.
Finally, in some instances properties that are not realised in the surface word form
are absent from the annotation
(e.g. in English the person and number of verbs are only annotated for 
third-person singular, as there are no distinct morphological forms for
their other values). 

\begin{figure*}
\includegraphics[width=\textwidth]{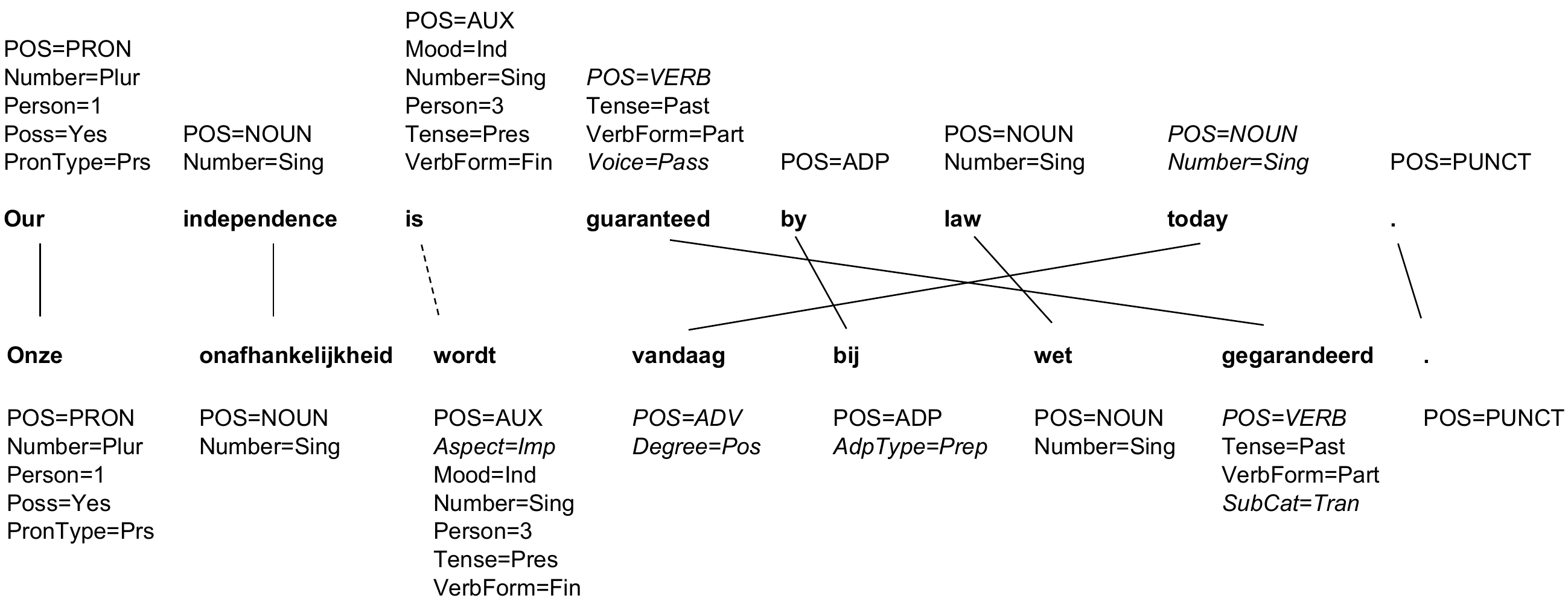}
\caption{A parallel sentence in English and Dutch annotated with universal 
morphological tags, showing high-confidence automatic word-alignments.
Attribute-value pairs that occur
only on one side of an aligned pair of tokens are indicated in \emph{italics}. The dashed
line indicates a low-confidence alignment point, which is ignored in our projection method.
}
\label{fig-ex}
\end{figure*}

An example of the morphological annotation employed is given in Figure 
\ref{fig-ex}.
Note that the annotations for aligned word-pairs are not fully consistent.
Some attributes appear only in the English treebank (e.g.\ Voice), while
others appear only in the Dutch treebank (e.g.\ Aspect, Subcat).

\section{Tag Projection across Bitext}

Our approach to train morphological taggers is based on the paradigm of
projecting token and type constraints as proposed by \newcite{TackstromDPMN13}.
The training data consist of parallel text with the resource-rich
language on the source-side and the low-resource language on the target side.
The source-side text is tagged with a supervised morphological tagger.
For every target-side sentence, the type and token constraints are used to
construct a set of permitted tags for each token in the sentence.
These constraints will then be used to train morphological taggers.

\subsection{Type and token constraints}

To extract constraints from the parallel text, we first obtain 
bidirectional word alignments. 
To ensure high quality alignments, alignment pairs with a confidence below a 
fixed threshold $\alpha$ are removed.
The motivation for using only high-confidence alignments is that incorrect
alignments will hurt the performance of the model, while it is easier to use
more parallel text to obtain a sufficient number of alignments for training.

The first class of constraints that we extract from the parallel text is
type constraints. 
For each word type, we construct a distribution over tags for the word by
accumulating counts of the morphological tags of source-side tokens 
that are aligned to instances of the word type.
The set of tags
with probability above some threshold $\beta$
is taken as the tag dictionary entry for that word type.
To construct the training examples, each token whose type occurs in the tag 
dictionary is restricted to the set of tags in the dictionary entry. 
For tokens for which the dictionary entry is empty, all the tags are included 
in the set of permitted tags (this happens when the
tag distribution is too flat and all the probabilities are below the threshold).
In principle, type constraints can also be obtained from an external 
dictionary, but
in this paper we assume we do not have such a resource.

The second class of constraints places restrictions on word tokens.
Every target token is constrained to the tag of its aligned source token,
while unaligned tokens can take any tag.

Token constraints are combined with type constraints as
proposed by \newcite{TackstromDPMN13}: 
If a token is unaligned, its type constraints are used.
If the token is aligned, and there is no dictionary entry for
the token type, the token constraint is used.
If there is a dictionary entry for the token type, and the token 
constraint tag is in the dictionary, the token constraint is used.
If the token constraint tag is not in the dictionary entry, the type
constraints are used.

\section{Learning from Projected Tags}

Next we propose models to learn a morphological tagger from cross-lingually
projected constraints.

\subsection{Related work}

HMMs have previously been used for weakly-supervised learning from 
token or type constraints~\cite{DasP11,LiGT12,TackstromDPMN13}. 
HMMs are generative models, and in this setting the words in the target 
sentence form the observed sequence and the morphological tags 
the hidden sequence. 
The projected constraints are used as partially observed training 
data for the hidden sequence.

\newcite{TackstromDPMN13} proposed a discriminative CRF model
that relies on incorporating two sets of constraints, of which one is a 
subset of the other. 
\newcite{GanchevD13} used a similar CRF model, but instead of using the 
projected tags as hard constraints, they were employed as soft constraints
with posterior regularization. 

The model of \newcite{WisniewskiPGY14} makes greedy predictions with
a history-based model, that includes previously
predicted tags in the sequence, during training and testing.
The model is trained with a variant of the perceptron algorithm that allows a 
set of positive labels. 
When an incorrect prediction is made during training,
the parameters are updated in the direction of all the positive labels.

\subsection{HMM model}

As a baseline model we use an HMM where
the transition and emission distributions are parameterized by log-linear
models (a feature-HMM). Training is performed with L-BFGS 
rather than with the EM algorithm. 
This parameterization was proposed by \newcite{BergKirkpatrickBDK10}
and applied to cross-lingual POS induction by \newcite{DasP11} and 
\newcite{TackstromDPMN13}.

Let $\mathbf{w}$ be the target sentence and $\mathbf{t}$ the
sequence of tags for the sentence.
The marginal probability of a sequence during training is
\[ p(\mathbf{w}_{1:n}) = \sum_{\mathbf{t}_{1:n} \in \mathcal{T}} 
               \prod_{i=1}^n p(t_i | t_{i-1}) p(w_i | t_i),
\]
where $\mathcal{T}$ is the set of tag sequences allowed by the type and 
token constraints. 
The probability of all other tag sequences are assumed to be $0$.

The features in our model are similar to those used by \newcite{TackstromDPMN13},
including features based on word and tag identity, suffixes up to length $3$,
punctuation and word clusters. 
Word clusters are obtained by clustering frequent words into 256 clusters with
the Exchange algorithm~\cite{UszkoreitB08}, using the data and methodology detailed
in \newcite{TackstromMU12}.

\subsection{Wsabie model}

We propose a discriminative model based on Wsabie~\cite{WestonBU11}, a shallow 
neural network that learns to optimize precision at the top of a ranked list
of labels.
In our application, the goal is to learn to rank the set of tags allowed
by the projected constraints in the training data above all other tags.
In contrast to the HMM, which performs inference over the entire sequence, 
Wsabie makes the predictions at each token independently, based on a large 
context-size.
Therefore, Wsabie inference is linear in the number of tags, while for an 
HMM it is quadratic, making the Wsabie model much faster during training and 
decoding.

Wsabie maps the input features and output labels into a low-dimensional joint space.
The input vector $x$ for a word $w$ consists of the concatenation of word
embeddings and sparse features extracted from $w$ and the surrounding context.
A mapping 
\[ \Theta_I(x) = Vx \]
maps $x \in \mathbb{R}^d$ into $\mathbb{R}^D$,
with matrix $V \in \mathbb{R}^{D \times d}$ of parameters.
The output tag $t$ is mapped into the same space by 
\[ \Theta_O(t) = W_t, \] 
where $W \in \mathbb{R}^{D \times L}$ is a matrix of output tag embeddings
and $W_t$ selects the column embedding of tag $t$.
The model score for tag $t$ given input token with feature vector $x$ is the dot product
\[f_t(x) = \Theta_O(t)^T \Theta_I(x), \] 
where the tags are ranked by the magnitude of $f_t(x)$. 
The norms of the columns of $V$ and $W$ are constrained,
which acts as a regularizer. 

The loss function is a margin-based hinge loss based on the rank of a tag 
given by $f_t(x)$. 
The rank is estimated by sampling an incorrect tag uniformly with replacement 
until the sampled tag violates the margin with a correct tag.
Training is performed with stochastic gradient descent by performing a 
gradient step against the violating tag.

The word embedding features for the Wsabie models consist of $64$-dimensional 
word vectors of the $5$ words on either side of a token and of the token itself.
The embeddings are trained with word2vec~\cite{MikolovSCCD13} on 
large corpora of newswire text.

Sparse features are based on prefixes and suffixes up to length $3$ as well as 
word cluster features for a window size $3$ around the token, using the clusters
described in the previous section.

\section{Experiments}

We evaluate our model in two settings. 
The first evaluation 
measures the accuracy of the cross-lingual taggers on language pairs 
where annotated data is available for both languages. 
The annotated target language data is used only during evaluation and not for training.
Second, we perform a downstream evaluation by including the morphological 
attributes predicted by the tagger as features in a dependency parser to
guage the effectiveness of our approach in a setting where one does not have access to
gold morphological annotations.

\subsection{Experimental setup}

As source of parallel training data we use 
Europarl\footnote{\url{http://www.statmt.org/europarl/}}~\cite{Koehn05} 
version 7. 
Sentences are tokenized but not lower-cased, and sentences longer than 80 
words are excluded. 
In our experiments we learn taggers for a set of 11 European languages that have both UD
training data with morphological features, and parallel data in Europarl:
Bulgarian, Czech, Danish, Dutch, Finnish, Italian, Polish, Portuguese, 
Slovene, Spanish and Swedish.
We train cross-lingual models in two setups:
The first uses English as source language;
in the second we train models with different source languages for
each target language.

Word alignments over the parallel data are obtained using 
FastAlign~\cite{DyerCS13}. 
High-confidence bidirectional word alignments are constructed by
intersecting the alignments in the two directions and including alignment
points only if the posterior probabilities in both directions are above the
alignment threshold $\alpha$. 
For each language pair all the word-aligned parallel data available
(between $10$ and $50$ million target-side tokens per language) are used
to extract the type constraints, and the models are trained on 
a subset of $2$ million target-side tokens (optionally with their token
constraints).

The number of distinct attribute-value pairs appearing in the tagsets
depends on the language pair and ranges
between $35$ and $79$, with $54$ on average (including POS tags).
The number of distinct composite morphological tags is $423$ on 
average, with a much larger range, between $81$ and $1483$. 
The English UD data has 116 tags composed out of 51 
distinct attribute-value pairs. 
Therefore, we can project a reasonable number of morpho-syntactic
attributes from
English, although the number of attribute combinations that occur
in the data is less than for morphologically richer languages.

The source text is tagged with supervised taggers, trained with Wsabie 
on the UD training data for each of the source languages used. 
For each language pair, we train a distinct source-side model covering 
only the attribute types appearing in both languages. 
This is meant to obtain a maximally accurate source-side tagger,
while accepting that our approach cannot predict target-side attributes
that are absent from the source language.
The average accuracy of the English taggers on the UD test data is $94.96\%$.
The source-side taggers over all the language pairs we experiment on have an
average accuracy of $95.75\%$, with a minimum of $89.14\%$ and a maximum of $98.59\%$.

\subsection{Tuning}

The hyperparameters of the Wsabie taggers are tuned on the English development 
set, and the same parameters are used for the Wsabie target-side models trained
on the projected tags.
The optimal setting is a learning rate of $0.01$, embedding dimension size $D=50$,
margin $0.1$, and 25 training iterations.

Hyperparameters for the projection models are set by tuning on the
UD dev set accuracy for English to Danish.
English was chosen as it is the language with the most available data and the most
likely to be used when projecting to other languages; Danish simply because 
its corpus size is typical of the larger languages in Europarl.
Using a small grid search, we choose the parameters that give the best average
accuracy across
all four projection model instances we consider.
This allows using the same hyperparameters for all these models,
an important factor in making them comparable in the evaluation, since
the hyperparameters determine the effective training data.
The parameters tuned in this manner are the alignment threshold $\alpha$, 
which is set to $0.8$, and the type distribution threshold $\beta$, set to $0.3$.

\subsection{Tagging evaluation setup}

In order to evaluate the induced taggers on the annotated UD data for the
target languages, we define two settings that circumvent mismatches between
source and target language annotations to different degrees.

The \textsc{Standard} setting involves first making minor corrections to certain
predicted POS values to account for inconsistencies in the original annotated
data.  When predicted by the model, the POS tag values absent from the target
language training corpus are deterministically mapped to the most-related value
present in the target language in the following way: \texttt{PROPN} to 
\texttt{NOUN}; \texttt{SYM} and \texttt{INTJ} to \texttt{X}; 
\texttt{SYM} and \texttt{X} to \texttt{PUNCT}. 
Besides POS, the evaluation considers only those
attribute \emph{types} that appear in both languages' training corpora, i.e.,
the set of attributes for which the model was trained. Note that this leaves
cases intact where the model predicts certain attribute \emph{values} that
appear only in one of the two languages; it is thus penalised for
making mistakes on values that it cannot learn under our projection approach.

The second evaluation setting, \textsc{Intersected}, relaxes the latter aspect:
it only considers attribute-value pairs appearing in the training corpora of
both languages.
The motivation for this is to get a better measurement of the 
accuracy of our method, assuming that the tagsets are consistent. 

In both settings we report macro-averaged F1 scores over all the considered
attribute types.
Results for Wsabie are averaged over 3 random restarts because it uses
stochastic optimization during training.

\subsection{Tagging results projecting from English}

\begin{table*}
\centering
\begin{tabular}{l|cc|cc|cc}
  Model & \multicolumn{2}{c|}{\textsc{Standard}} & \multicolumn{2}{c|}{\textsc{Intersected}} & \multicolumn{2}{c}{POS} \\
\hline
HMM projected type & 53.86 & (-) & 58.67 & (-) & 79.45 & (-) \\
HMM projected type and token & 48.49 & (-) & 52.40 & (-) & 73.61 & (-) \\
\hline
unambiguous type & 51.72 & (0.33) & 56.22 & (0.36) & 79.58 & (0.22) \\
projected type & 53.60 & (0.16) & 58.11 & (0.18) & 80.09 & (0.12) \\
projected type and token & 53.36 & (0.19) & 57.77 & (0.21) & 79.94 & (0.11) \\
\hline
  supervised 1K    & 62.44 & (1.52) & 61.74 & (1.55) & 72.51 & (0.82) \\
\hline
  supervised type  & 75.55 & (1.88) & 74.72 & (1.95) & 75.91 & (1.16) 
\end{tabular}
\caption{Cross-lingual morphological tagging from English: Macro F1 scores averaged across 
$11$ languages. 
All the results except for the first two rows are for Wsabie models.
The standard deviation over $3$ runs is given in brackets.
}
\label{tab-all-results}
\end{table*}

Following previous work on projecting POS tags
and the assumption that it is easier to obtain parallel data between 
a low-resource language and English than with another language,
we start by training cross-lingual taggers using English as source language.

The overall tagging results are given in Table \ref{tab-all-results}.
In addition to evaluating the morphological tags in the two settings
described above, we also report accuracies for POS tags only, projected jointly
with the morphological attributes. 

We find that for both the HMM and Wsabie models the performance with 
type and token constraints is worse than when only using type constraints.
\newcite{TackstromDPMN13} similarly found that for HMMs for POS projection,
models with joint constraints do not perform better than those using only 
type constraints.
They postulated that this is due to the type dictionaries having the same
biases as token projections, and therefore the model with joint constraints 
not being able to filter out systematic errors in the projections.

For both sets of constraints the performance of the Wsabie model is close
to that of the corresponding HMM, despite the Wsabie model having a linear
runtime against the quadratic runtime of the HMM.

As another baseline we train a Wsabie model on unambiguous type 
constraints, i.e., we only extract training examples for words which only have
a single tag in the tag dictionary.
Including ambiguous type constraints gives an average improvement of $2.2\%$.

As a target ceiling on performance we train a Wsabie model with supervised type 
constraints.
This model uses type constraints based on an oracle morphological tag 
dictionary extracted from the gold training data of the target language. 
It is trained on the same training data as the projected models (without token
constraints).
The model scores higher on \textsc{Standard} than on \textsc{Intersected},
as it has access to annotations for the full set of tags used in the target 
language, not just the restricted set that can be projected.
This oracle performs on average $17\%$ better than the projected type
constraints model on \textsc{Intersected}.
Therefore, despite the promising results of our approach, there is still a 
considerable amount of noise in the type constraints extracted from the aligned
data.

We also compare the performance of the model to that of a supervised model 
trained on a small annotated corpus. 
Average performance when training on $1000$ annotated tokens is only a few
points higher than that of the best projected model for \textsc{Intersected}.
Given that is it expensive to let annotators learn to annotate a large set of
attributes, even for a small corpus, it shows that our model can 
bring considerable benefits in practice to the development of NLP models for 
low-resource languages.
It is possible to obtain further improvements in performance by learning 
jointly from a small annotated dataset and parallel 
data~\cite{DuongEa14}, but we leave that for future work.

The results when evaluating only the POS tags follow the same pattern, 
except that the overall level of accuracy is much higher than when considering all
morphological attributes. 
For POS, the models with projected constraints actually perform better than
those with supervised type constraints.
In this case the benefits from learning constraints from a larger set of word 
types seem to outweigh the noise in the projections.
The projected models are also more accurate than the supervised model trained
on $1000$ tokens.

\subsection{Multilingual tagging results}

\begin{table*}
\small
\centering
\begin{tabular}{l|ccccccccccc|c}
   & bg   & cs   & da   & es  & fi   & it  & nl  & pl   & pt   & sl   & sv & \bf{Avg.} \\
\hline
  en & 46.7 & 49.7 & 58.0 & 55.7 & 54.0 & 59.6 & \bf{64.1} & 45.0 & 57.8 & 51.0 & 47.9 & 53.6 \\
\hline
bg & - & 58.3 & 59.2 & 51.2 & 52.6 & 43.2 & 38.7 & 52.8 & 41.1 & 49.2 & 53.6 & 50.0 \\
  cs & 55.2 & - & 54.5 & 42.3 & 48.4 & 51.3 & 45.0 & \bf{56.8} & 33.6 & \bf{67.5} & 53.2 & 50.8 \\
  da & \bf{61.9} & \bf{61.6} & - & 41.8 & 49.1 & 45.5 & 49.6 & 53.7 & 44.0 & 49.3 & \bf{72.1} & 52.9 \\
  es & 54.3 & 58.8 & 41.3 & - & 53.0 & \bf{74.4} & 52.1 & 52.2 & \bf{69.2} & 53.8 & 46.9 & 55.6 \\
fi & 46.6 & 48.7 & 45.3 & 39.5 & - & 50.9 & 36.8 & 37.4 & 30.1 & 55.5 & 57.8 & 44.9 \\
  it & 43.6 & 59.4 & 44.0 & \bf{74.0} & 53.3 & - & 54.3 & 46.5 & 69.2 & 55.9 & 47.0 & 54.7 \\
nl & 44.7 & 59.5 & 56.2 & 54.8 & 54.0 & 60.3 & - & 55.9 & 58.6 & 48.6 & 51.6 & 54.4 \\
pl & 52.7 & 58.6 & 46.3 & 37.5 & 42.1 & 47.9 & 42.1 & - & 40.7 & 56.0 & 42.6 & 46.6 \\
pt & 45.4 & 45.0 & 49.6 & 66.2 & 42.6 & 69.5 & 50.1 & 43.5 & - & 47.8 & 43.9 & 50.3 \\
sl & 46.6 & 60.7 & 35.2 & 40.9 & 49.2 & 49.8 & 36.0 & 54.1 & 35.0 & - & 40.4 & 44.8 \\
  sv & 50.1 & 54.6 & \bf{70.7} & 47.7 & \bf{57.2} & 49.7 & 46.9 & 41.6 & 46.3 & 43.5 & - & 50.8 \\
\hline
  \bf{Avg} & 49.8 & 55.9 & 50.9 & 50.1 & 50.5 & 54.7 & 46.9 & 49.0 & 47.8 & 52.6 & 50.6 &
\end{tabular}
  \caption{Cross-lingual morphological tagging results (\textsc{Standard} F1 scores) per source and target language,
Wsabie projected model with type constraints. Rows indicate source language and columns target language.
}
\label{tab-multilingual-tag-results}
\end{table*}

Results for cross-lingual experiments on all pairs of the target languages under consideration
are given in Table \ref{tab-multilingual-tag-results}, using the \textsc{Standard} evaluation setup.
We make use of Wsabie for these experiments, as it is a more efficient model, which is especially 
significant when training models with large tagsets.

We see that there is large variance in the morphological tagging accuracies across language pairs.
In most cases the source language for which we learn the most accurate model for morphological tagging on the 
target language is a related language.
The Romance languages we consider (Spanish, Italian and Portuguese) seem to transfer particularly well across each other.
Swedish and Danish also transfer well to each other, while English transfers best to Dutch, which the former is most closely related
to among the languages compared here.
However, there are also some cases of unrelated source languages performing best: 
Using Danish as source language gives the highest performing models for both Bulgarian and Czech.
When comparing these results, however, one should keep in mind that the attribute type sets used to train
taggers from different source languages for the same target language is not always the same 
(due to our definition of the \textsc{Standard} evaluation), therefore these results should not be
interpreted directly as indicating which source language gives the best target language performance on 
a particular tagset.

\begin{table}
\centering
\begin{tabular}{l|cc|cc}
  & \multicolumn{2}{c|}{\small \textsc{Standard} } & \multicolumn{2}{c} {\small \textsc{Intersected} } \\
  & en- & best- & en- & best- \\ 
\hline
bg & 46.7 & 61.88 & 51.6 & 64.97 \\
cs & 49.7 & 61.57 & 55.7 & 63.97 \\
da & 58.0 & 70.74 & 65.4 & 73.14 \\
es & 55.7 & 74.01 & 60.7 & 74.62 \\
fi & 54.0 & 57.23 & 59.1 & 59.11 \\
it & 59.6 & 74.42 & 66.1 & 75.32 \\
nl & 64.1 & 64.12 & 64.7 & 64.66 \\
pl & 45.0 & 56.83 & 47.3 & 60.39 \\
pt & 57.8 & 69.22 & 60.2 & 73.10 \\
sl & 51.0 & 67.48 & 53.4 & 69.86 \\
sv & 47.9 & 72.07 & 55.1 & 74.60 
\end{tabular}
\caption{Comparison of the performance of the most accurate cross-lingual taggers
  for each target language, compared to having English as  source language.} 
\label{tab-lang}
\end{table}

We compare the results of the \textsc{Standard} and \textsc{Intersected} evaluations,
both when using English as source language, and when using the source language which gives 
the highest accuracy
on \textsc{Standard} for each target language (Table \ref{tab-lang}).
We see that the gap in performance between the two evaluations tends to be larger
when projecting from English than when projecting from the source language which performs 
best for each target language. 

One of the main causes of variation in performance is annotation differences.
Languages that are morphologically rich tend to have lower performance, 
but we also see variation between similar languages: There is a $10\%$ performance
gap between Danish and Swedish when projecting from English, even though they are closely related.

\begin{table}
\centering
\begin{tabular}{l|cc}
Target & en- & best- \\ 
\hline
  bg & 81.84 & 81.84 (en) \\ 
  cs & 80.41 & 86.29 (sl) \\ 
  da & 80.69 & 84.85 (sv) \\ 
  es & 86.02 & 89.04 (it) \\ 
  fi & 77.07 & 77.48 (cs) \\  
  it & 83.46 & 86.91 (es) \\
  nl & 73.05 & 76.02 (da) \\ 
  pl & 79.38 & 82.66 (cs) \\  
  pt & 84.30 & 87.98 (es) \\ 
  sl & 74.71 & 83.21 (cs) \\ 
  sv & 80.37 & 86.47 (da) \\ 
\end{tabular}
\caption{Wsabie projected model with type constraints, POS accuracy with
English and the best language for each target as source.
}
\label{tab-pos}
\end{table}

We also investigate the effect of the choice of source language on the accuracy of the projected POS
tags (Table \ref{tab-pos}).
Again, we compare the performance with English as source (which is standard for previous
work on POS projection) to that of the best source language for each target.
Although the gap in performance is smaller than for the full evaluation, 
we see that for most target languages we can still do better by projecting from a language
other than English.

\begin{table*}
\small
\centering
\addtolength{\tabcolsep}{-1.1pt} 
\begin{tabular}{l|cc|cc|cc|cc|cc|cc|cc|cc|cc|cc|cc}
Target  & \multicolumn{2}{c|}{bg} &	\multicolumn{2}{c|}{cs} & \multicolumn{2}{c|}{da} & \multicolumn{2}{c|}{es} & \multicolumn{2}{c|}{fi} & \multicolumn{2}{c|}{it} & \multicolumn{2}{c|}{nl} & \multicolumn{2}{c|}{pl} & \multicolumn{2}{c|}{pt} & \multicolumn{2}{c|}{sl} & \multicolumn{2}{c}{sv} \\ 
\hline
  Source & en & da & en & it & en & sv & en & it & en & sv & en & pt & en & en & en & nl & en & it & en & cs & en & da \\ 
\hline
  Case   &	  40 & 62  &	2 & -  & 62 & 18  &	4 & - &	5 & 26  &- & -  &	16 & 16  &	4 & 4  &	50 & -  &	2 & 68  &	10 & 14 \\ 
  Definite   & 1 & 68 &	- & -  &	0 & 64  &	97 & 97  &	- & -  &	89 & 91  &	89 & 89  &	- & -  &	93 & 93  &	2 & -  &	19 & 66 \\ 
  Degree   &  67 & 63 &	69 & 2  &	72 & 77  &	5 & 26  &	50 & 47  &	2 & 14  &	56 & 56  &	57 & 47  &	2 & 18 &	63 & 74  & 70 & 81  \\ 
  Gender   &	 1 & 6 &	2 & 46  &	7 & 78  &	0 & 85  &	- & -  &	2 & 80  &	0 & 0  &	3 & 0  &	1 & 77  &	2 & 61  &	7 & 81  \\ 
  Mood   &	 61 & 66 &	55 & 83 &	81 & 94  &	72 & 80  &	69 & 79  &	76 & 83  &	69 & 69  &	58 & 63  &	74 & 75  & 68 & 91  &	73 & 94 \\ 
  Number   &	69 & 71  &	67 & 75  &	60 & 82 &	54 & 92  &	67 & 68  &	57 & 90  &	78 & 78  &	69 & 63  &	62 & 75  &	68 & 91  &	64 & 94 \\ 
  NumType   &	 64 & 62  &	91 & 86  &	84 & -  &	82 & 85 &	86 & - &	89 & 66 &	78 & 78  &	- & -  &	63 & 65  & 86 & 73  &	- & - \\ 
  Person   &	 54 & 28  &	63 & 68  &	56 & -  &	58 & 79  &	51 & -  &	57 & 80  &	82 & 82  &	55 & 59  &	61 & 74  & 67 & 91  &	- & -   \\ 
  Poss   &	 76 & 77  &	90 & 84  &	97 & 98  &	94 & 93  &	- & -  &	87 & 88  &	64 & 64  &	- & -  &	96 & 98  &	67 & 62  & 99 & 97  \\ 
  PronType   & 72 & 71  &	41 & 38 &	46 & 2  &	82 & 74  &	42 & 0  &	76 & 71 &	81 & 81  &	38 & 50  &	81 & 79  & 43 & 73  &	0 & 3  \\ 
  Reflex   &	 0 & 85  &	0 & 0  &	62 & - &	0 & 0 &	61 & -  &	0 & 0  &	60 & 60  &	0 & 97  &	0 & 0  &	- & - &	- & -  \\ 
  Tense   &	 60 & 63  &	68 & 70  &	81 & 85  &	69 & 81  &	67 & 77  &	75 & 75  &	74 & 74  &	66 & 66  &	64 & 72  & 62 & 74  &	65 & 86 \\ 
  VerbForm   & 43 & 49  &	59 & 78  &	75 & 79  &	79 & 81  &	64 & 65  &	82 & 81  &	78 & 78  &	56 & 66  &	79 & 72  & 59 & 74  &	73 & 86 \\ 
  Voice   & 9 & 75  &	9 & - &	6 & 89  &	- & - &	10 & 76  &	- & -  &	- & -  &	55 & -  &	- & -  &	- & -  &	15 & 90 \\  
\end{tabular}
\addtolength{\tabcolsep}{1.1pt} 
\caption{Cross-lingual tagging results (F1 scores) per language and per 
  attribute (not showing POS and a small number of attribute types that only appear with 1 or 2 language pairs), for Wsabie projected with type constraints.
  English and best source language.
}
\label{tab-attr-results}
\end{table*}

Detailed per attribute results for the \textsc{Standard} evaluation are given in 
Table~\ref{tab-attr-results}, again comparing the results of projecting from English
to that of the most accurate model for each target language.
We see that there are large differences in accuracy across attributes and
across languages.
In some cases, the transfer is unsuccessful.
For example, degree accuracy in Italian is $2\%$ F1 when projecting from English and
$14\%$ F1 projecting from  Portuguese.
Some of the cases can be explained by differences in where an attribute is
marked: 
For example, for definiteness the performance is $1\%$ from English to
Bulgarian, as Bulgarian marks definiteness on nouns and adjectives rather than
on determiners.
Other attributes are very language-dependent.
Gender transfers well between Romance languages, but poorly when
transferring from English.

\subsection{Parsing evaluation}

To evaluate the effect of our models on a downstream task, we apply the 
cross-lingual taggers induced using English as source language to dependency parsing.
This is applicable to a scenario where a language might have a corpus annotated with
dependency trees and universal POS, but not morphological attributes.
We want to determine how much of the performance gain from features
based on supervised morphological tags we can recover with the tags predicted 
by our model. 

\begin{table}
\centering
\begin{tabular}{l|c|c|c}
& no morph  & projected type & supervised \\ 
\hline
bg & 79.14 & 78.99 & 79.62  \\
cs & 76.88 & 77.25 & 79.03  \\ 
da & 69.73 & 70.04 & 71.51  \\
es & 77.66 & 78.08 & 78.64  \\ 
fi & 61.78 & 62.68 & 70.42  \\
it & 81.51 & 81.49 & 82.24  \\
nl & 64.76 & 65.80 & 65.92  \\
pl & 70.83 & 71.89 & 74.03  \\
pt & 75.92 & 76.71 & 77.98  \\
sl & 77.17 & 77.46 & 79.25  \\
sv & 72.92 & 74.09 & 74.58  \\
\hline
\textbf{Avg.} & 73.48 & 74.04 & 75.75 
\end{tabular}
\caption{Dependency parsing results (LAS) with no, projected and supervised 
morphological tags.}
\label{tab-parse}
\end{table}

As baseline we use a reimplementation of \newcite{ZhangN11}, 
an arc-eager transition-based dependency parser with a rich feature-set, 
with beam-size $8$, trained for $10$ epochs with a structured perceptron.	
We assume that universal POS tags are available, using a supervised SVM POS
tagger for training and evaluation.

To include the morphology, we add features
based on the predicted tags of the word on top of the stack and the first two words on
the buffer.

Parsing results are given in Table \ref{tab-parse}. 
We report labelled attachment scores (LAS) for the baseline with no
morphological tags, the model with features predicted by Wsabie with 
projected type constraints, and the model with features predicted 
by the supervised morphological tagger.

We obtain improvements in parsing accuracies for all languages except 
Bulgarian when adding the induced morphological tags.
Using the projected tags as features recovers $24.67\%$ ($0.6$ LAS absolute) of
the average gain that supervised morphology features delivers over the baseline
parser.
The parser with features from the supervised tagger trained on $1000$ tokens
obtains $73.63$ LAS on average.
This improvement of +$0.15$ LAS over the baseline versus the +$0.6$ of our method
shows that the tags predicted by our projected models are more useful as
features than those predicted by a small supervised model.

To investigate the effect of source language choice for the projected models in this 
evaluation, we trained a model for Swedish using Danish as source language.
The parsing performance is insignificantly different from using 
English as source, despite the accuracy of the tags projected from Danish being
higher.

\newcite{FaruquiMS16} show that features from induced morpho-syntactic lexicons 
can also improve dependency parsing accuracy. 
However, their method relies on having a seed lexicon of $1000$ annotated word types,
while our method does not require any morphological 
annotations in the target language.

\section{Future Work}

A big challenge in cross-lingual morphology is that of relatedness between 
source and target languages.
Although we evaluate our models on multiple source-target language pairs, more
work is required to investigate strategies for choosing which source language to
use for a low-resource target language.
A related direction is to constructing models from multiple source languages, 
as our results show that the overall best-performing source language for a given
target language may not always have the best performance on all attributes.

Another direction is to make use of dictionaries such as Wiktionary to obtain
type constraints, similar to previous work on weakly-supervised POS 
tagging~\cite{LiGT12,TackstromDPMN13}. 
\newcite{SylakGlassmanKYQ15} and \newcite{SylakGlassmanKPQY15} proposed a 
morphological schema and method to extract annotations in that schema from 
Wiktionary. 
Although different from the schema used in this paper, their method can be used
to extract type dictionaries for morphological tags that can be used to 
complement constraints extracted from parallel data.

Finally, greater use can be made of syntactic information: There is a close
relation between the syntactic structure expressed in dependency parses
and inflections in morphologically rich languages; by including this 
syntactic structure in our models we can induce morphological tags, e.g. related 
to case, that is also expressed in dependency parses.

\section{Conclusion}

In this paper we proposed a method that can successfully induce morphological
taggers for resource-scarce languages using tags projected across bitext.
It relies on access to a morphological tagger for a source-language and a moderate
amount of bitext.
The method obtains strong performance on a range of language pairs.
We showed that downstream tasks such as dependency parsing can be improved by 
using the predictions from the tagger as features.
Our results provide a strong baseline for future work in weakly-supervised
morphological tagging.

\section*{Acknowledgments}

This research was primarily performed while the first author was an intern at Google Inc. 
We thank Oscar T\"ackstr\"om, Kuzman Ganchev, Bernd Bohnet and Ryan McDonald 
for valuable assistance and discussions about this work.

\bibliography{acl2016}

\begin{thebibliography}{}

\bibitem[\protect\citename{Ahlberg \bgroup et al.\egroup }2015]{AhlbergFH15}
Malin Ahlberg, Markus Forsberg, and Mans Hulden.
\newblock 2015.
\newblock Paradigm classification in supervised learning of morphology.
\newblock In {\em Proceedings of NAACL}, pages 1024--1029.

\bibitem[\protect\citename{Berg-Kirkpatrick \bgroup et al.\egroup
  }2010]{BergKirkpatrickBDK10}
Taylor Berg-Kirkpatrick, Alexandre Bouchard-C{\^o}t{\'e}, John DeNero, and Dan
  Klein.
\newblock 2010.
\newblock Painless unsupervised learning with features.
\newblock In {\em Proceedings of NAACL}, pages 582--590.

\bibitem[\protect\citename{Besacier \bgroup et al.\egroup }2014]{BesacierBKS14}
Lauent Besacier, Ettiene Barnard, Alexey Karpov, and Tanja Schultz.
\newblock 2014.
\newblock Automatic speech recognition for under-resourced languages: A survey.
\newblock {\em Speech Communication}, 56:85--100.

\bibitem[\protect\citename{Creutz and Lagus}2007]{CreutzL07}
Mathias Creutz and Krista Lagus.
\newblock 2007.
\newblock Unsupervised models for morpheme segmentation and morphology
  learning.
\newblock {\em ACM Transactions on Speech and Language Processing}, 4(1).

\bibitem[\protect\citename{Das and Petrov}2011]{DasP11}
Dipanjan Das and Slav Petrov.
\newblock 2011.
\newblock Unsupervised part-of-speech tagging with bilingual graph-based
  projections.
\newblock In {\em Proceedings of ACL}, pages 600--609.

\bibitem[\protect\citename{de Marneffe \bgroup et al.\egroup
  }2014]{MarneffeEa14}
Marie-Catherine de~Marneffe, Timothy Dozat, Natalia Silveira, Katri Haverinen,
  Filip Ginter, Joakim Nivre, and Christopher~D. Manning.
\newblock 2014.
\newblock Universal dependencies: A cross-linguistic typology.
\newblock In {\em Proceedings of LREC}.

\bibitem[\protect\citename{Duong \bgroup et al.\egroup }2014]{DuongEa14}
Long Duong, Trevor Cohn, Karin Verspoor, Steven Bird, and Paul Cook.
\newblock 2014.
\newblock What can we get from 1000 tokens? {A} case study of multilingual pos
  tagging for resource-poor languages.
\newblock In {\em Proceedings of EMNLP}, pages 886--897.

\bibitem[\protect\citename{Durrett and DeNero}2013]{DurrettD13}
Greg Durrett and John DeNero.
\newblock 2013.
\newblock Supervised learning of complete morphological paradigms.
\newblock In {\em Proceedings of NAACL}, pages 1185--1195.

\bibitem[\protect\citename{Dyer \bgroup et al.\egroup }2013]{DyerCS13}
Chris Dyer, Victor Chahuneau, and Noah~A Smith.
\newblock 2013.
\newblock A simple, fast, and effective reparameterization of {IBM} {M}odel 2.
\newblock In {\em Proceeding of NAACL}, pages 682--686.

\bibitem[\protect\citename{Faruqui \bgroup et al.\egroup }2016]{FaruquiMS16}
Manaal Faruqui, Ryan McDonald, and Radu Soricut.
\newblock 2016.
\newblock Morpho-syntactic lexicon generation using graph-based semi-supervised
  learning.
\newblock {\em Transactions of the Association for Computational Linguistics},
  4:1--16.

\bibitem[\protect\citename{Fossum and Abney}2005]{FossumA05}
Victoria Fossum and Steven Abney.
\newblock 2005.
\newblock Automatically inducing a part-of-speech tagger by projecting from
  multiple source languages across aligned corpora.
\newblock In {\em Proceedings of IJCNLP}, pages 862--873.

\bibitem[\protect\citename{Ganchev and Das}2013]{GanchevD13}
Kuzman Ganchev and Dipanjan Das.
\newblock 2013.
\newblock Cross-lingual discriminative learning of sequence models with
  posterior regularization.
\newblock In {\em Proceedings of EMNLP}, pages 1996--2006.

\bibitem[\protect\citename{Haji{\v{c}} \bgroup et al.\egroup }2009]{HajicEa09}
Jan Haji{\v{c}}, Massimiliano Ciaramita, Richard Johansson, Daisuke Kawahara,
  Maria~Ant{\`o}nia Mart{\'\i}, Llu{\'\i}s M{\`a}rquez, Adam Meyers, Joakim
  Nivre, Sebastian Pad{\'o}, Jan {\v{S}}t{\v{e}}p{\'a}nek, et~al.
\newblock 2009.
\newblock The {CoNLL}-2009 shared task: Syntactic and semantic dependencies in
  multiple languages.
\newblock In {\em Proceedings of CoNLL: Shared Task}, pages 1--18.

\bibitem[\protect\citename{Koehn}2005]{Koehn05}
Philipp Koehn.
\newblock 2005.
\newblock Europarl: A parallel corpus for statistical machine translation.
\newblock In {\em MT summit}, volume~5, pages 79--86.

\bibitem[\protect\citename{Li \bgroup et al.\egroup }2012]{LiGT12}
Shen Li, Jo\~{a}o Gra\c{c}a, and Ben Taskar.
\newblock 2012.
\newblock Wiki-ly supervised part-of-speech tagging.
\newblock In {\em Proceedings of EMNLP-CoNLL}, pages 1389--1398, July.

\bibitem[\protect\citename{Mikolov \bgroup et al.\egroup }2013]{MikolovSCCD13}
Tomas Mikolov, Ilya Sutskever, Kai Chen, Greg~S Corrado, and Jeff Dean.
\newblock 2013.
\newblock Distributed representations of words and phrases and their
  compositionality.
\newblock In {\em Advances in neural information processing systems}, pages
  3111--3119.

\bibitem[\protect\citename{M\"{u}ller and Schuetze}2015]{MullerS15}
Thomas M\"{u}ller and Hinrich Schuetze.
\newblock 2015.
\newblock Robust morphological tagging with word representations.
\newblock In {\em Proceedings of NAACL}, pages 526--536, Denver, Colorado,
  May--June.

\bibitem[\protect\citename{Nivre \bgroup et al.\egroup }2015]{NivreEa15}
Joakim Nivre, {\v Z}eljko Agi{\'c}, Maria~Jesus Aranzabe, Masayuki Asahara,
  Aitziber Atutxa, Miguel Ballesteros, John Bauer, Kepa Bengoetxea, Riyaz~Ahmad
  Bhat, Cristina Bosco, Sam Bowman, Giuseppe G.~A. Celano, Miriam Connor,
  Marie-Catherine de~Marneffe, Arantza Diaz~de Ilarraza, Kaja Dobrovoljc,
  Timothy Dozat, Toma{\v z} Erjavec, Rich{\'a}rd Farkas, Jennifer Foster,
  Daniel Galbraith, Filip Ginter, Iakes Goenaga, Koldo Gojenola, Yoav Goldberg,
  Berta Gonzales, Bruno Guillaume, Jan Haji{\v c}, Dag Haug, Radu Ion, Elena
  Irimia, Anders Johannsen, Hiroshi Kanayama, Jenna Kanerva, Simon Krek,
  Veronika Laippala, Alessandro Lenci, Nikola Ljube{\v s}i{\'c}, Teresa Lynn,
  Christopher Manning, C{\v a}t{\v a}lina M{\v a}r{\v a}nduc, David Mare{\v
  c}ek, H{\'e}ctor Mart{\'i}nez~Alonso, Jan Ma{\v s}ek, Yuji Matsumoto, Ryan
  {McDonald}, Anna Missil{\"a}, Verginica Mititelu, Yusuke Miyao, Simonetta
  Montemagni, Shunsuke Mori, Hanna Nurmi, Petya Osenova, Lilja {\O}vrelid,
  Elena Pascual, Marco Passarotti, Cenel-Augusto Perez, Slav Petrov, Jussi
  Piitulainen, Barbara Plank, Martin Popel, Prokopis Prokopidis, Sampo Pyysalo,
  Loganathan Ramasamy, Rudolf Rosa, Shadi Saleh, Sebastian Schuster, Wolfgang
  Seeker, Mojgan Seraji, Natalia Silveira, Maria Simi, Radu Simionescu, Katalin
  Simk{\'o}, Kiril Simov, Aaron Smith, Jan {\v S}t{\v e}p{\'a}nek, Alane Suhr,
  Zsolt Sz{\'a}nt{\'o}, Takaaki Tanaka, Reut Tsarfaty, Sumire Uematsu, Larraitz
  Uria, Viktor Varga, Veronika Vincze, Zden{\v e}k {\v Z}abokrtsk{\'y}, Daniel
  Zeman, and Hanzhi Zhu.
\newblock 2015.
\newblock Universal dependencies 1.2.
\newblock {LINDAT}/{CLARIN} digital library at Institute of Formal and Applied
  Linguistics, Charles University in Prague.

\bibitem[\protect\citename{Petrov \bgroup et al.\egroup }2012]{PetrovDM12}
Slav Petrov, Dipanjan Das, and Ryan McDonald.
\newblock 2012.
\newblock A universal part-of-speech tagset.
\newblock In {\em Proceedings of LREC}.

\bibitem[\protect\citename{Snyder and Barzilay}2008]{SnyderB08}
Benjamin Snyder and Regina Barzilay.
\newblock 2008.
\newblock Unsupervised multilingual learning for morphological segmentation.
\newblock In {\em Proceedings of ACL}, pages 737--745.

\bibitem[\protect\citename{Sylak-Glassman \bgroup et al.\egroup
  }2015a]{SylakGlassmanKPQY15}
John Sylak-Glassman, Christo Kirov, Matt Post, Roger Que, and David Yarowsky.
\newblock 2015a.
\newblock A universal feature schema for rich morphological annotation and
  fine-grained cross-lingual part-of-speech tagging.
\newblock In {\em Proceedings of Systems and Frameworks for Computational
  Morphology: Fourth International Workshop}, pages 72--93. Springer
  International Publishing, Cham.

\bibitem[\protect\citename{Sylak-Glassman \bgroup et al.\egroup
  }2015b]{SylakGlassmanKYQ15}
John Sylak-Glassman, Christo Kirov, David Yarowsky, and Roger Que.
\newblock 2015b.
\newblock A language-independent feature schema for inflectional morphology.
\newblock In {\em Proceedings of ACL-IJCNLP (short papers)}, pages 674--680.

\bibitem[\protect\citename{T{\"a}ckstr{\"o}m \bgroup et al.\egroup
  }2012]{TackstromMU12}
Oscar T{\"a}ckstr{\"o}m, Ryan McDonald, and Jakob Uszkoreit.
\newblock 2012.
\newblock Cross-lingual word clusters for direct transfer of linguistic
  structure.
\newblock In {\em Proceedings of NAACL}, pages 477--487.

\bibitem[\protect\citename{T{\"a}ckstr{\"o}m \bgroup et al.\egroup
  }2013]{TackstromDPMN13}
Oscar T{\"a}ckstr{\"o}m, Dipanjan Das, Slav Petrov, Ryan McDonald, and Joakim
  Nivre.
\newblock 2013.
\newblock Token and type constraints for cross-lingual part-of-speech tagging.
\newblock {\em Transactions of the Association for Computational Linguistics},
  1:1--12.

\bibitem[\protect\citename{Toutanova \bgroup et al.\egroup
  }2008]{ToutanovaSR08}
Kristina Toutanova, Hisami Suzuki, and Achim Ruopp.
\newblock 2008.
\newblock Applying morphology generation models to machine translation.
\newblock In {\em Proceedings of ACL-HLT}, pages 558--566.

\bibitem[\protect\citename{Tsarfaty \bgroup et al.\egroup }2010]{TsarfatyEa10}
Reut Tsarfaty, Djam{\'e} Seddah, Yoav Goldberg, Sandra K{\"u}bler, Marie
  Candito, Jennifer Foster, Yannick Versley, Ines Rehbein, and Lamia Tounsi.
\newblock 2010.
\newblock Statistical parsing of morphologically rich languages ({SPMRL}):
  what, how and whither.
\newblock In {\em Proceedings of the NAACL HLT 2010 First Workshop on
  Statistical Parsing of Morphologically-Rich Languages}, pages 1--12.

\bibitem[\protect\citename{Uszkoreit and Brants}2008]{UszkoreitB08}
Jakob Uszkoreit and Thorsten Brants.
\newblock 2008.
\newblock Distributed word clustering for large scale class-based language
  modeling in machine translation.
\newblock In {\em Proceedings of ACL-HLT}, pages 755--762.

\bibitem[\protect\citename{Weston \bgroup et al.\egroup }2011]{WestonBU11}
Jason Weston, Samy Bengio, and Nicolas Usunier.
\newblock 2011.
\newblock Wsabie: Scaling up to large vocabulary image annotation.
\newblock In {\em Proceedings of the International Joint Conference on
  Artificial Intelligence (IJCAI)}.

\bibitem[\protect\citename{Wisniewski \bgroup et al.\egroup
  }2014]{WisniewskiPGY14}
Guillaume Wisniewski, Nicolas Pécheux, Souhir Gahbiche-Braham, and François
  Yvon.
\newblock 2014.
\newblock Cross-lingual part-of-speech tagging through ambiguous learning.
\newblock In {\em Proceedings of EMNLP}, pages 1779--1785.

\bibitem[\protect\citename{Yarowsky \bgroup et al.\egroup }2001]{YarowskyNW01}
David Yarowsky, Grace Ngai, and Richard Wicentowski.
\newblock 2001.
\newblock Incuding multilingual text analysis tools via robust projection
  across aligned corpora.
\newblock In {\em Proceedings of HLT}.

\bibitem[\protect\citename{Zeman}2008]{Zeman08}
Daniel Zeman.
\newblock 2008.
\newblock Reusable tagset conversion using tagset drivers.
\newblock In {\em Proceedings of LREC}.

\bibitem[\protect\citename{Zhang and Nivre}2011]{ZhangN11}
Yue Zhang and Joakim Nivre.
\newblock 2011.
\newblock Transition-based dependency parsing with rich non-local features.
\newblock In {\em Proceedings of ACL-HLT}, pages 188--193.

\end{thebibliography}
\bibliographystyle{acl2016}

\end{document}